# Constructing Parallel Multidimensional Chromatic Lexicons for Corpus-Assisted Analysis of Russian and English Texts

Larisa Nikitina
Faculty of Languages and Linguistics
Universiti Malaya
Malaysia

## Abstract

This article addresses the relative scarcity of research tools for the corpus-assisted linguistic analysis of colour terms in literary texts. It describes the development of two multidimensional chromatic lexicons: one for Russian (224 entries) and one for English (141 entries). Lexicon construction involved sourcing colour vocabulary from specialised resources and research literature, comparing the two language inventories, manually checking translated candidates, and addressing language-specific morphological features. In addition to identifying colour terms and visual descriptors, the lexicons classify entries according to hue, saturation, and temperature. To demonstrate their practical application, a pilot study was conducted on purposively sampled corpora of poetry by Andrei Bely (20,373 tokens) and Emily Dickinson (28,479 tokens). All retrieved matches were checked in context and classified as Confirmed_chromatic, Ambiguous_visual, or Excluded. The analysis was implemented in two main stages: a strict analysis including confirmed chromatic lexis only, followed by a sensitivity analysis incorporating both confirmed and ambiguous chromatic lexis to determine whether coding decisions about borderline cases affected the main findings. The quantitative results indicated marked differences in the use of colour terms, visual descriptors, hue, saturation, and temperature. Specifically, the analysis revealed that confirmed chromatic terms occurred 3.4 times more frequently in the sampled Bely corpus than in the Dickinson corpus. These findings demonstrate the analytical value of a multidimensional approach, with the main contribution of this study being a transparent and reusable procedure for constructing and applying multilingual chromatic lexicons.

## Introduction

Colours, or more precisely human perceptions of them, trigger a wide range of physiological, psychological, and cognitive responses. The affective and cognitive functions of colours are well recognised, and these qualities are harnessed across various domains of human activity. For instance, commercial brands carefully consider colour palettes in their logos. Aware that colour schemes influence perception, politically conservative parties tend to avoid using red, as it is traditionally associated with revolutionary movements; instead, they favour darker hues in their logos and banners. Even brief phrases containing a colour term, such as All Blacks or Manchester Red, elicit a wide range of emotions from sports fans. Likewise, the slogan “Go Green” is universally understood as a call for environmental awareness.

Given this profound sensory impact of colours on humans, extensive research on chromatic perceptions and descriptions has been conducted across various academic disciplines (Fox, 2023). A major impetus for systematic scientific research on the use and evolution of colour terms came from Berlin and Kay’s (1969/1991) seminal study, in which they collected extensive data on the colour lexicons of 20 world languages. They identified what constitutes a basic colour term and established 11 basic colour terms (BCTs) in English: *black, white, red,* *yellow, green, blue, pink, purple, brown, grey,* and *orange*. Furthermore, Berlin and Kay proposed that the basic colour terms in languages other than English align with these 11 colour categories. This proposition implies that chromatic lexis reflects universal perceptual categories. They also argued that colour lexicons are not static and that the evolution of colour terms in any language follows a hierarchical and predictable sequence. Empirical studies have subsequently tested the veracity of these propositions. While many studies have offered empirical support for Berlin and Kay's hypothesis (McManus, 1983), some scholars have expressed reservations regarding the universality of the colour lexicon (e.g., Wierzbicka, 2008).

In applied linguistics, studies examining colour lexicons and cross-linguistic differences in colour naming have a long history, spanning over 160 years. The colour lexicon is defined by Lindsey and Brown (2021, p. 606) as "the collection of words used within a language community to name colour categories." Referencing Linhares et al. (2008), Lindsey and Brown pointed out that humans can distinguish about $10^6$ distinct colours, whereas colour lexicons in human languages are limited to the order of $10^1$. In addition to lexicon composition, scholars have investigated the structure and historical development of semantic categories for colour names (Biggam, 2012; Bochkarev et al., 2023), colour perception and communication (Griber et al., 2021), and individual or cultural influences on colour perception (Golda et al., 2022; Griber et al., 2021; Lindsey & Brown, 2014). Some studies have also examined the associations between colour terms and human emotions (Mohr et al., 2018). The next section offers a more extensive review of the literature.

## Research on Colour Terms in the Social Sciences and Humanities

The pragmatics and semantics of colour terms used in works of literature have attracted considerable interest from researchers. One of the earliest studies by Pratt (1898, as cited in McManus, 1983) examined the use of colour terms in a large corpus of 17 English poets, from John Gower (1330–1408) to John Keats (1795–1821). Pratt found that the most frequently used colour terms in the corpus of poetic works were *red* and *green*.

Adopting a literary linguistics perspective, researchers have explored colour imagery and the functions of colour terms in folk songs and in the idiolects of poets writing across various epochs, languages, and genres (e.g., Giorgianni, 2020; İstanbullu, 2023; Özşahin, 2023; Salih Hassan, 2000; Ševčíková, 2008). References to colours are frequently found in folk songs. For example, Özşahin (2023) identified numerous colour references in Bashkir bride ballads, while İstanbullu (2023) found that Turkish folk song lyrics often use colour terms to express personal feelings, beliefs, and hopes, or to heighten the vividness of social and narrative descriptions. Giorgianni (2020) explored the use of colour terms in references to love and sensuality in ancient Greek poetry and in texts written by physicians from the same historical period. The researcher noted a tendency among poets and physicians in antiquity to employ different colours to portray different facets of love. Green, for example, was used to allude to love as a disease, and red to denote love as a fire that ignites passion and consumes a person. The Greek term manifold colours was employed to evoke the deceptive nature of love.

Among poets writing in English, William Shakespeare (1564–1616) and Emily Dickinson (1830–1886) are notable for their creative use of colour terms and rich colour palettes. Referencing earlier studies, Fox (2023, p. 30) reported that in a corpus of over 800,000 words from William Shakespeare's works, there are approximately 800 colour terms. *Black* accounts for almost a quarter of Shakespeare's colour idiolect (202 instances), followed by *white* (169), *green* (116), and *red* (107). Ševčíková (2008), in her analysis of colour terms in Emily Dickinson's poetry, reported that the poet used approximately 50 different colours. Among them, purple appeared in 54 poems and was the most frequently mentioned colour. As Salih Hassan (2000) argued, Dickinson employed references to purple to form a deeper spiritual connection with the inner self and to express all that is precious in life. Other colours frequently used by Dickinson include basic hues such as *blue*, *white*, and *yellow*. Notably, the poet's rich colour idiolect included references to comparatively rare terms, such as *cochineal* and *indigo*.

Some studies have adopted a semantic linguistics perspective to investigate how basic colour terms are used in different cultural and linguistic contexts. Golda et al. (2022) explored the use of colour terms and their cultural connotations in English, French, Italian, Polish, and Japanese contexts. The researchers discovered both universally shared features and cultural differences in references to basic colours. For example, in European languages and in Japanese, references to green relate to nature, plants, and vegetation. More broadly, green is often equated with vegetation and ecology. The sun is associated with the colour yellow in all the languages studied. These are universally shared perceptions of green and yellow, as reported in many studies (Fox, 2023). Inevitably, as noted by Golda et al. (2022), there are some cultural differences in colour perceptions, even in culturally similar contexts. An interesting difference among European languages concerns references to the colour blue. In French, this colour is often used in idiomatic expressions to refer to anger, whereas in other European languages, anger is typically associated with red (Golda et al., 2022).

More recently, the linguistic representation of colour terms in song lyrics has become the subject of scholarly research. Umam and Nirmala (2020) explored and compared the use of colour metaphors in English and Indonesian songs, focusing on seven colours: *blue, grey, yellow, white, red, green,* and *black*. The researchers discovered striking similarities in the connotative functions of colour terms in English and Indonesian. Both languages associated *black* with darkness, *white* with purity, *yellow* with enlightenment, and *grey* with sadness. At the same time, Umam and Nirmala pointed out significant differences in the use of *red* and *green*. Specifically, in the sampled English song lyrics, red represented love and passion, while in Indonesian songs, it was associated with resistance, fury, and bravery. Regarding *green*, Umam and Nirmala concluded that this colour was frequently associated with money in English songs, whereas in Indonesian song lyrics, it represented fertility and the environment.

There is a growing recognition that people's responses to physical colours, as well as their linguistic references to colours, can convey a wide range of emotions. These perceptions can be highly heterogeneous, even within the same cultural context. Referencing a number of earlier studies, Mohr et al. (2018) pointed out that red, the most studied colour in research literature, has been associated with a wide range of negative emotions, including anger and danger. At the same time, red has also been used to convey happiness, elation, cheerfulness, love, and passion. The literature shows that colour terms can perform salient emotional, symbolic, and stylistic functions in literary and other artistic texts. However, the interpretation of an individual occurrence remains context-dependent: a colour expression may be descriptive, metaphorical, conventional, or polysemous. Existing studies are frequently author-specific or restricted to basic colour terms, and reusable cross-linguistic resources for systematic corpus-assisted analysis remain limited. A further methodological problem is that colour vocabulary is often treated as a homogeneous category. This overlooks distinctions among hue, saturation, temperature, and visually relevant descriptors. The present study therefore develops two aligned but language-specific multidimensional lexicons. The term aligned is used deliberately: the resources share a common analytical schema, but they are not assumed to be translation-equivalent inventories. The study addresses the following research questions:

RQ1. What methodological principles support the construction of aligned chromatic lexicons for morphologically distinct languages such as Russian and English?

RQ2. How can hue, saturation, temperature, and descriptor categories be operationalised consistently while preserving language-specific lexical distinctions?

RQ3. What corpus-specific patterns do the lexicons identify in the sampled poetry of Andrei Bely and Emily Dickinson, and how stable are those patterns under strict and inclusive coding decisions?

The study's primary outputs are two machine-readable lexicons intended for corpus stylistics and other types of corpus-assisted analyses. A small pilot study comparing two poets is included as an illustrative application of how the lexical categories can generate systematic, replicable findings. It also shows how ambiguous colour-related terms can be handled through sensitivity analysis.

## Methodology for Constructing a Multidimensional Chromatic Lexicon

### Sourcing the Colour Terms

The development of a reliable lexicon follows a structured sequence or set of phases (Drake & Civille, 2003; Nikitina, 2024). The phases involved in constructing the colour lexicons are outlined in Table 1 and explained in detail below. To compile colour terms effectively and ensure adequate coverage and accuracy, authoritative sources must be consulted. Because the study of colour is a highly specialised area, a comprehensive inventory of colour terms far exceeds the 11 basic colours (BCTs).

In the initial step, specialised online resources listing colour terms were identified. Candidate English terms were collected from ColourHexa (https://www.colourhexa.com/), whereas candidate Russian terms were sourced from ColourScheme.ru (https://colourscheme.ru/colour-names.html). Although these resources provide extensive lists and visual colour representations, their entries were treated as provisional data and were not accepted automatically. The initial inventories were expanded and verified using the LIWC-22 Dictionary Workbench function (Pennebaker et al., 2022), and then cross-checked against relevant scholarship on Russian and English colour terminology (Bochkarev et al., 2023; Griber et al., 2018, 2021; Lindsey & Brown, 2014). Every retained lexical item was then manually evaluated for its suitability.

Following this initial compilation, two dedicated data files were created to organise the lexical inventories. In this step, the two inventories were alphabetised, deduplicated, and compared cross-linguistically. After cleaning and organising the data, the English colour terms were machine-translated into Russian and the Russian terms into English using Google Translate. Machine translation was used purely as a discovery aid to identify possible additions and omissions, rather than to establish semantic equivalence. Suggested additions were assessed manually for conventional usage and colour meaning in the target language. Terms lacking an appropriate equivalent in the target language were excluded.

**Table 1**

Stages in the construction of the English and Russian chromatic lexicons

| Stage | Action and Rationale | Outcome / Example |
|---|---|---|
| 1. Initial Sourcing | Chromatic terms were collected from online and scholarly sources: (1) ColourHexa (https://www.colourhexa.com/) for English, (2) ColourScheme.ru (https://colourscheme.ru/colour-names.html) for Russian, and (3) studies by Bochkarev et al. (2023), Griber et al. (2018, 2021), and Lindsey and Brown (2014). | Initial wordlists of English and Russian colour terms were compiled. |
| 2. Cross-Pollination | Each language-specific list was compared with machine-translated candidates from the other language to identify possible omissions and additions; translation was used as a discovery aid, not as evidence of equivalence. | Candidate counterparts were added only after manual assessment (e.g., English *lilac* prompted by Russian сиреневый [*sirenevyy*]). |
| 3. Manual Curation | Translated candidates were manually checked to remove terms that did not have an equivalent in the other language. Conventional gemstone/material colour terms were retained where appropriate; ambiguous corpus occurrences were separated during contextual coding. | Non-equivalent items were excluded, while potentially chromatic gem/material terms were handled through strict and sensitivity coding (e.g., unlike in English, the Russian term цитрин [*tsitrin*] refers to the gemstone only, and конфетный [*konfetnyy*] has no colour sense). |
| 4. Morphological Considerations | To ensure retrieval precision in Russian, a morphologically rich language, ambiguous stems were avoided. | For example, the stem желт (*zhelt-*) was excluded to prevent false matches with non-colour words such as желток (*zheltok*, egg yolk). |
| 5. Multidimensional Categorization | Retained entries were classified by hue, saturation, temperature, and term type using one shared operational schema. | The resulting categories enable comparable but language-sensitive corpus analysis. |
| 6. Inclusion of Descriptor Terms | Additional lexical items describing chromatically relevant aesthetic or visual properties were incorporated to capture subtle expressive variations. | Examples include the English *opaque, glowing, pale, vivid,* and the Russian светло- ([*svetlo-*] light-), тёмный/тёмно- [*tyomnyy*/*tyomno-*, dark-]. |
| 7. Outcome | Corrected lexicons were stored as tool-agnostic spreadsheets with documented coding rules and versioned corrections. | Russian resource: 224 entries; English resource: 141 entries (supplied as separate Supplementary Materials). |

## Cross-Linguistic Pollination of the Colour Term Lists

In this phase, an aligned lexical resource was developed and omissions were identified in each language-specific inventory. Alignment refers to the application of a shared schema and comparable analytical categories, rather than one-to-one translation equivalence. For example, the Russian сиреневый (*sirenevyy*, lilac) prompted the addition of *lilac* to the English candidate list. (Note: Russian terms in this article are transliterated using the BGN/PCGN 1947 system.)

During the cross-pollination process, it was essential to ensure that machine-translated terms were appropriate and semantically valid in both languages. Because colour terminology is inherently language-specific, providing direct translation matches was frequently challenging. Therefore, manual curation became critical; this process is described in the following section.

**Manual Check of Machine-Translated Terms**

During the manual curation stage, literal translation of colour terms and artificial lexical symmetries produced by machine translation that do not exist in natural language were removed. For example, the English *citrine* can denote both a gemstone and a colour, whereas Russian цитрин (*tsitrin*) denotes only the gemstone; similarly, English *candy* can function as a colour modifier, but Russian конфетный (*konfetnyy*) does not serve this function. Conversely, the Russian грушевый (*grushevyy*, pear-coloured) has no direct equivalent colour term in English.

Lexical items derived from gemstone or material sources that conventionally denote colour were retained. However, during corpus analysis, occurrences whose chromatic sense could not be unambiguously verified were coded as ambiguous rather than automatically counted as confirmed colour references. Following manual curation, the lexicon entries were structured within a shared machine-readable framework. Each entry was coded as a colour term or visual descriptor and classified according to hue, saturation, and temperature. The same analytical schema was applied to both Russian and English, while individual classifications reflected conventional meanings and morphological characteristics within each respective language.

**Morphological Considerations**

To ensure analytical precision, morphological complexity must be accounted for, particularly in highly inflected languages such as Russian. A salient feature of Russian is its extensive system of adjectival modifiers. For example, beyond the basic colour term зелёный (zelënyy, *green*), Russian employs prefixed modifiers to indicate shade, hue, or degree of *green*, such as зелёно- (*zelyono*-) or зеленовато- (*zelenovato*-) denoting *green* or *greenish*. This morphological productivity enables descriptions such as зелёно-серый (*zelëno-seryy,* grey-green) and зеленовато-желтый (*zelenovato-zhëltyy,* greenish-yellow). In contrast, English relies primarily on suffixes (e.g., *-ish*) or compound modifiers. The availability of these productive modifiers across many hues accounts for the substantially larger inventory in the Russian colour lexicon compared with English. However, such derivation is not applicable across all Russian colour terms. For instance, terms such as пурпуроватый (*purpurovatyy,* purplish) and пурпуровато- (*purpurovato*-) are not in standard use.

Where consistent with Russian grammatical norms, selected lexical entries were represented by stems. For example, *желтоват-* (*zheltovat-*, yellowish) captures related adjectival and adverbial variants. Broad stems, however, were avoided for most Russian entries to prevent retrieving non-chromatic terms. For instance, the stem *желт-* (zholt-) would match not only *жёлтый* (*zholtyy,* yellow) but also words such as *желток* (*zheltok*, egg yolk). To mitigate such false matches in the pilot study, the corpora were fully lemmatised prior to analysis. The software therefore matched exact dictionary base forms rather than relying on truncated wildcard stems.

Additional challenges in constructing the Russian lexicon stemmed from the morphological richness of the language and the rapid emergence of chromatic neologisms. As Griber et al. (2021) noted, loanwords and calques are ubiquitous in contemporary Russian colour vocabulary. Examples include лайм (*laym, lime*), шампань (*shampan'*, champagne), and капучино (*kapuchino, cappuccino*). Furthermore, some neologisms take the form of noun-based adjectives or nouns that serve a descriptive function (e.g., тыква [*tykva*, pumpkin colour]). For this reason, highly idiosyncratic neologisms listed on the *ColourScheme.ru* were excluded as they lack habitual usage. An example of an excluded term is зеленые джунгли крайола (*zelyonye dzhungli krayola*, green jungle crayola). Nevertheless, the Russian lexicon includes well-established colloquial expressions, such as серобуромалиновый (*sero-buro-malinovyy*), a humorous designation for an ill-defined, typically unpleasant shade (literally, grey-brown-raspberry).

To address potential issues arising from difficulties in recognising and categorising colour neologisms during computer-assisted analysis, the words цвет (*tsvet*, colour) and цвета (*tsveta*, colours), together with their English equivalents, were included in the colour lexicons. The proximity of these terms may signal that a reference to colour is being made in the text. This allows the analysis to flag the lexical item as belonging to the colour domain rather than to other semantic fields, such as food, beverages, flora, or fauna.

## Classification of the Colour Terms: Theoretical, Methodological, and Practical Considerations

The decision to categorise chromatic terms across the dimensions of hue, intensity (saturation), and temperature is theoretically grounded in established models of colour perception and cognitive psychology. In visual semantics and colorimetry, multidimensional frameworks such as the Munsell colour system (Munsell, 1912) and the HSV (Hue, Saturation, Value) colour model demonstrate that colour is not perceived and processed as a one-dimensional sensory phenomenon. Rather, colour perception involves a number of distinct processes. Furthermore, psycholinguistic and psychological research suggests that colour characteristics can initiate distinct affective and cognitive responses. For example, a study by Valdez and Mehrabian (1994) found that colour saturation (intensity) was a primary driver of emotional arousal; highly saturated colours consistently elicited stronger affective responses. Similarly, Ou et al. (2004) reported that categorising colours by temperature (warm vs. cool) provides a robust psychological construct associated with emotional valence, spatial perception, and physiological reactions. By embedding these perceptual dimensions into a structured language resource, the resulting lexicon affords analyses beyond basic frequency calculations: it enables researchers to systematically quantify patterns in an author's chromatic repertoire within a corpus under study.

Methodologically, the workflow adopted in this study follows the procedure described in a study on sensory lexicon construction (Nikitina, 2024). To enable a more nuanced linguistic analysis of colour terms, each entry was classified according to its hue, saturation level (intensity), and temperature (see Supplementary Files S1 and S2). The Hue category includes the labels *red, orange, yellow, green, blue, purple, brown, pink*, and *grey*. This classification represents conventional linguistic and perceptual colour categories.

Saturation was operationalised using the same rule in both lexicons: High denotes vivid, pure, or intense chromatic prototypes (e.g., *crimson*); *Medium denotes unmarked, mixed, or source-derived colour prototypes (e.g., apricot*, lavender); and Low denotes achromatic, pale, muted, greyed, or attenuated prototypes (e.g., *champagne*). Temperature was coded as Cool, Warm, and Neutral. To elaborate, cool colours are typically associated with a soothing and calming effect; they include *blue, green*, or combinations of these. Warm colours are considered energising and uplifting; they feature *orange* or *red* hues, or a combination of these. Neutral colours are not conventionally associated with either warmth or coolness and include *white, black, beige, tan,* and *grey*.

## Colour Descriptor Terms

The chromatic lexicons presented in this study include a Descriptor category. Its inclusion is informed by a recent study by Alvarado et al. (2024), who demonstrated that sensory concepts do not exist in isolation but are intricately connected with emotional and aesthetic descriptors in language. The category includes modifiers relevant to colour research: terms that do not name a specific hue, such as *yellow*, but describe visual appearance or aesthetic qualities. In English, these descriptors include adjectives such as *opaque*, *bright*, and *dark*. In Russian, the relevant adjectives may also occur in adverbial forms, for example, яркий (*yarkiy*) / ярко- (*yarko*-), светлый (*svetlyy*) / светло- (*svetlo*-), and тёмный (*tyomnyy*) / тёмно- (*tyomno*-), respectively.

Colour descriptors may be general (e.g., *opaque*), relate to light effects (e.g., *glowing, gleaming, flashing*), indicate colour saturation (e.g., *pale, vivid*), or describe the visual effect of colour (e.g., *harsh, soft*). Including these items as a separate analytical category makes it possible to conduct a more detailed linguistic analysis of chromatic expression. Importantly, descriptor terms may not have a vision-related meaning in every context, such as in "vivid memories" or "opaque procedure". A descriptor term is counted only when it refers to vision-related phenomena. The next section presents the results of a pilot study using the colour lexicons developed in this study.

## Illustrative Case Study: Bely and Dickinson

### Corpora and Analytical Procedure

To illustrate the analytical capabilities and practical application of the chromatic lexicons, a pilot study was conducted on a small sample of poems by Andrei Bely (1880–1934) and Emily Dickinson (1830–1886). While recognising that the two poets come from different historical periods and literary traditions, their selection was a deliberate methodological choice for maximum variation sampling. Both poets are noted for their extensive use of colour imagery (Fishman, 1988; Ševčíková, 2008); however, Bely is one of the leading Russian Symbolist poets, while Dickinson's work is situated within the tradition of American Romanticism. By testing the lexicons against these divergent chromatic idiolects, this study aims to evaluate their robustness and analytical capacity across different literary traditions. It should be noted that because the pilot study uses illustrative samples rather than complete collections of the poets' work, it does not aim to make general claims about the use of colour terms in Russian and English poetry.

The texts were prepared for analysis by converting them to lowercase, removing punctuation and non-textual material (e.g., years when poems were written), and lemmatising the words. After cleaning and lemmatisation, the Bely corpus contained 20,373 tokens and the Dickinson corpus contained 28,479 tokens. For the analysis, a table was prepared in which each lexical item identified by the lexicons in each corpus was classified as *Confirmed_chromatic* (terms that clearly refer to colour, light, or visible appearance, e.g., blue sky), *Ambiguous_visual* (mentions of gemstones, metals, or expressions whose colour meaning could not be established with certainty, e.g., gold, silver, ruby), or *Excluded* (terms that did not refer to colour or any visual phenomenon). Because the two corpora differed in size, both raw frequencies and normalised frequencies per 10,000 tokens were calculated.

For a more robust analysis, two statistical analyses were conducted. First, the strict analysis included *Confirmed_chromatic* terms only; second, the sensitivity analysis included both *Confirmed_chromatic* and *Ambiguous_visual* occurrences. These two separate tests were conducted to assess whether the inclusion of borderline cases in the sensitivity test affected the statistical results.

The log-likelihood statistic ($G^2$) assessed whether the frequency differences in the use of colour words across the two corpora were statistically significant (Dunning, 1993). Log ratios were computed to assess effect sizes. Because some terms occurred with low frequencies (e.g., the hues *orange, brown*, and *pink*, and the *medium* saturation in the strict analysis, as well as the hues *orange* and *pink* in the sensitivity analysis), Fisher's exact test was performed in both the strict and sensitivity analyses for these cases. Furthermore, to reduce the risk of Type I errors arising from multiple comparisons, the unadjusted $G^2$ *p*-values were corrected using the Holm sequential procedure (Holm, 1979). The analyses were implemented in R using four scripts. The scripts used to calculate Fisher's exact test results, log-likelihood statistics and *p* values, Holm-adjusted *p* values, and log ratios are provided as Supplementary Files S3–S6, respectively.

**Findings from the Strict Analysis**

Table 2 shows the full results of the strict analysis. The strict analysis identified 304 confirmed chromatic occurrences in the Bely corpus (149.22 per 10,000 tokens) and 125 in the Dickinson corpus (43.89 per 10,000 tokens). This difference was statistically significant, $G^2 = 148.96$, Holm-adjusted $p < .001$. The log ratio of 1.76 indicates that confirmed chromatic expressions occurred approximately 3.4 times more frequently in the sampled Bely corpus. The Bely corpus also contained considerably more visual descriptors, including words referring to brightness, darkness, radiance, and visual intensity (78 occurrences, 38.29 per 10,000 tokens), compared with the Dickinson corpus (10 occurrences, 3.51 per 10,000 tokens), $G^2 = 84.91$, Holm-adjusted $p < .001$.

**Table 2**

Comparison of confirmed chromatic expressions in the Bely and Ddickinson corpora

| Category | Bely, *n* (per 10,000) | Dickinson, *n* (per 10,000) | G² | Holm-adjusted *p* | Log ratio |
|---|---|---|---|---|---|
| **Overall** | | | | | |
| All chromatic expressions | 304 (149.22) | 125 (43.89) | 148.96 | < .001 | 1.76 |
| Colour terms | 226 (110.93) | 115 (40.38) | 83.50 | < .001 | 1.45 |
| Visual descriptors | 78 (38.29) | 10 (3.51) | 84.91 | < .001 | 3.39 |
| **Hue** | | | | | |
| Red | 67 (32.89) | 28 (9.83) | 32.21 | < .001 | 1.73 |
| Orange | 15 (7.36) | 0 (0.00) | 26.24 | < .001 | 5.44 |
| Yellow | 17 (8.34) | 11 (3.86) | 4.09 | .173 | 1.09 |
| Green | 22 (10.80) | 6 (2.11) | 15.86 | < .001 | 2.27 |
| Blue | 44 (21.60) | 17 (5.97) | 23.12 | < .001 | 1.83 |
| Purple | 7 (3.44) | 21 (7.37) | 3.42 | .193 | −1.04 |

| Category | Bely, *n* (per 10,000) | Dickinson, *n* (per 10,000) | G² | Holm-adjusted *p* | Log ratio |
|---|---|---|---|---|---|
| Brown | 1 (0.49) | 5 (1.76) | 1.74 | .263 | −1.39 |
| Pink | 6 (2.95) | 3 (1.05) | 2.28 | .263 | 1.38 |
| Grey | 21 (10.31) | 2 (0.70) | 25.30 | < .001 | 3.59 |
| **Saturation** | | | | | |
| High | 146 (71.66) | 87 (30.55) | 41.37 | < .001 | 1.23 |
| Medium | 10 (4.91) | 0 (0.00) | 17.49 | < .001 | 4.88 |
| Low | 113 (55.47) | 24 (8.43) | 96.42 | < .001 | 2.70 |
| **Temperature** | | | | | |
| Cool | 51 (25.03) | 23 (8.08) | 22.31 | < .001 | 1.62 |
| Warm | 74 (36.32) | 65 (22.82) | 7.48 | .006 | 0.67 |
| Neutral | 70 (34.36) | 19 (6.67) | 50.65 | < .001 | 2.34 |

*Note.* Bely corpus = 20,373 tokens; Dickinson corpus = 28,479 tokens. Positive log ratios indicate a higher normalised frequency in Bely; negative values indicate a higher frequency in Dickinson. Log ratios were calculated using a 0.5 continuity correction. Holm correction was applied separately to the Overall, Hue, Saturation, and Temperature families. Because some lexical entries received more than one classification, category frequencies should not be added together. Fisher's exact tests supported the conclusions obtained from the sparse comparisons that *orange* ($p < .001$) and *medium* saturation ($p < .001$) were significant, whereas *brown* ($p = .411$) and *pink* ($p = .177$) were not significant.

At the level of individual hues, red, orange, green, blue, and grey were significantly more frequent in Bely after Holm correction. Although yellow was also more frequent in Bely, this difference was not statistically significant. The differences for purple, brown, and pink were likewise non-significant. All three saturation categories and all three temperature categories occurred significantly more frequently in Bely. Fisher's exact tests supported the conclusions obtained from the sparse comparisons that orange ($p < .001$) and medium saturation ($p < .001$) were significant, whereas brown ($p = .411$) and pink ($p = .177$) were not significant.

### Sensitivity Analysis

The sensitivity analysis examined whether the findings were robust when terms classified as ambiguous visual expressions (e.g., *gold*, *silver*, and *ruby*) were included. Their inclusion increased the overall normalised frequency to 211.55 occurrences per 10,000 tokens in Bely ($n = 431$) and 58.64 in Dickinson ($n = 167$). The difference remained statistically significant, $G^2 = 225.79$, Holm-adjusted $p < .001$, with a log ratio of 1.85, which indicates that these expressions occurred approximately 3.6 times more frequently in the sampled Bely corpus. Thus, the overall difference between the two corpora was not an artefact of how ambiguous cases were classified.

**Table 3**

Selected results of the sensitivity analysis including ambiguous visual expressions

| Category | Bely, *n* (per 10,000) | Dickinson, *n* (per 10,000) | $G^2$ | Holm-adjusted *p* | Log ratio |
|---|---|---|---|---|---|
| All chromatic and ambiguous visual expressions | 431 (211.55) | 167 (58.64) | 225.79 | < .001 | 1.85 |
| Yellow | 90 (44.18) | 26 (9.13) | 62.04 | < .001 | 2.26 |

*Note.* Bely corpus = 20,373 tokens; Dickinson corpus = 28,479 tokens. Log ratios were calculated using a 0.5 continuity correction. Holm correction was applied within the relevant families of comparisons. The main shift in results concerned the hue yellow. Its frequency difference was not significant in the strict analysis ($p = .173$), but became statistically significant when gold and related material-source expressions were included ($G^2 = 62.04$, Holm-adjusted $p < .001$). This finding demonstrates that the classification of material-source terms can alter conclusions regarding specific hues, even as the overall comparative baseline between the corpora remains stable.

## Discussion

This study demonstrates how parallel colour lexicons can be developed and used to analyse and compare chromatic language in Russian and English texts. The main outcome of this study is not merely the statistical divergence between Bely and Dickinson, but the methodological demonstration of a multidimensional language resource for the analysis of colour terms. The procedure follows a staged approach to sensory lexicon construction (Nikitina, 2024). However, the present study takes this procedure further by adapting it to the particular complexity of colour vocabulary. Its main advances are the classification of chromatic terms across several dimensions, the alignment of the two lexicons with attention to their different morphological systems, and the inclusion of visual descriptors as a distinct analytical category.

With regard to the first research question, the process of constructing the two lexicons shows that cross-linguistic alignment cannot be achieved by simply translating one list of colour terms into another. Russian and English differ considerably in how colour meanings are expressed, particularly because Russian colour vocabulary permits extensive derivation and the formation of compound and modified colour terms. The two lexicons were therefore made comparable through a shared analytical structure, while the selection and morphological representation of individual terms remained language-specific. The second research question concerned whether hue, saturation, temperature, and descriptor categories could be applied consistently while at the same time preserving these language-specific differences. The findings suggest that this is possible when the categories are defined through common operational rules but applied with reference to the conventional meaning of each lexical item.

The pilot analysis illustrates the value of these methodological decisions. Confirmed chromatic expressions occurred approximately 3.4 times more frequently in the sampled Bely corpus than in the Dickinson corpus. The categories also revealed differences in the poets' use of visual descriptors, individual hues, saturation levels, and colour temperatures. The overall difference remained stable when ambiguous expressions were added, although the finding for *yellow* changed. This shows why borderline expressions should not simply be included or excluded without further examination: their classification may not affect the overall pattern, but it can alter the interpretation of a particular chromatic category.

The findings from applying the lexicons to a sample corpus provide a quantitative framework for comparison with previous qualitative research on colour use. Ševčíková (2008) examined the symbolism of white in Dickinson's poetry; Salih Hassan (2000) identified purple as a particularly prominent colour in the American poet's work and explored its associations with spirituality, time, and eternity. While purple was more frequent in Dickinson's poems than in Bely's in the present sample, which aligns with previously reported findings, it was not the dominant hue in Dickinson's sampled corpus. This result does not necessarily contradict Salih Hassan's findings because the studies differ in corpus coverage, classification procedures, and analytical purpose. Rather, it highlights an additional value of the systematic, replicable workflow afforded by using the lexicon: a colour's symbolic importance and its statistical frequency are not equivalent constructs.

Quantitative results obtained by using the colour lexicons as a research tool can suggest promising directions for further qualitative explorations of how colour expressions are employed within literary works, and how they define an author's colour idiolect. For example, while the use of yellow in Bely's work is well established in the literature (Fishman, 1988), the high frequency of low-saturation terms and visual descriptors identified in this study suggests that further exploration could focus on how luminosity, obscurity, and chromatic intensity contribute to Bely's poetic imagery. Also, the lexicons could be used to examine colour terms in various other artistic texts, such as song lyrics (Liu & Nikitina, 2025), or to explore patterns of colour usage in marketing and product descriptions. The lexicons may also be applied to larger digital datasets, including online consumer reviews. In these contexts, lexicon-based findings could help identify recurrent patterns, generate hypotheses, and triangulate qualitative interpretations.

While this pilot study has yielded several notable insights, its findings must be interpreted within certain limitations. It examines only one poet per language, and the samples vary by author, language, historical period, and literary tradition. The results therefore describe these two sampled corpora and should not be generalised to Russian and English poetry as a whole.

## Conclusion, Limitations, and Future Directions

This study presents machine-readable chromatic lexicons containing 224 Russian and 141 English entries. The difference in their size stems from the greater morphological and derivational productivity of the Russian colour vocabulary. By categorising colour terms and visual descriptors according to hue, saturation, and temperature, the lexicons offer a structured resource for analysing the use of chromatic terms in various artistic and non-literary texts. Several limitations must be acknowledged. Although the Russian and English lexicons offered in this study are extensive, they are not exhaustive. Furthermore, some entries, particularly source-derived colour terms and visual descriptors, are polysemous, and their occurrences (e.g., bronze sky vs. bronze statue) must be manually verified and interpreted within context. In addition, the Bely and Dickinson corpora are purposive samples rather than complete authorial corpora. Despite these unavoidable limitations, the present study provides a much-needed practical language resource and demonstrates how a multidimensional classification of colour terms can enable systematic quantitative analysis while retaining the contextual judgement required for literary texts.

## References

Alvarado, J. A., Velasco, C., & Salgado, A. (2024). The organization of semantic associations between senses in language. *Language and Cognition, 16*(4), 1588–1617. https://doi.org/10.1017/langcog.2024.19

Berlin, B., & Kay, P. (1991). *Basic color terms: Their universality and evolution.* University of California Press. (Original work published 1969)

Biggam, C. P. (2012). *The semantics of colour: A historical approach.* Cambridge University Press.

Bochkarev, V. V., Shevlyakova, A. V., Solovyev, V. D., Rakhilina, E. V., & Paramei, G. V. (2023). Linguistic mechanisms of colour term evolution: A diachronic investigation of "Russian browns" *buryj* and *koričnevyj*. *Diachronica, 40*(4), 492–531. https://doi.org/10.1075/dia.22013.boc

Dickinson, E. (2004). *Poems by Emily Dickinson: Three series, complete.* Project Gutenberg. https://www.gutenberg.org/ebooks/12242

Drake, M. A., & Civille, G. V. (2003). Flavor lexicons. *Comprehensive Reviews in Food Science and Food Safety, 2*(1), 33–40. https://doi.org/10.1111/j.1541-4337.2003.tb00013.x

Dunning, T. (1993). Accurate methods for the statistics of surprise and coincidence. *Computational Linguistics, 19*(1), 61–74.

Fishman, L. (1988). *Symbolism of colours in Bely's "Petersburg"* [Master's thesis, University of Manitoba]. MSpace. https://mspace.lib.umanitoba.ca/bitstreams/7cbd247d-22cc-485b-8b6a-3d70fdee008b/download

Fox, J. (2023). *The world according to colour: A cultural history.* Penguin Books.

Giorgianni, F. (2020). Colori dell'eros nella Grecia antica [Colours of love in ancient Greece]. *Medicina nei Secoli: Arte e Scienza, 32*(2), 443–476.

Golda, P., Jedziniak, A., Mężyk, J., Ryszka, J., & Uchman, T. (2022). Colour terms in five linguistic images of the world: The semantic perspective. *GEMA Online Journal of Language Studies, 22*(4), 39–58. https://doi.org/10.17576/gema-2022-2204-03

Griber, Y. A., Mylonas, D., & Paramei, G. V. (2018). Objects as culture-specific referents of colour terms in Russian. *Color Research & Application, 43*(6), 958–975. https://doi.org/10.1002/col.22280

Griber, Y. A., Mylonas, D., & Paramei, G. V. (2021). Intergenerational differences in Russian color naming in the globalized era: Linguistic analysis. *Humanities and Social Sciences Communications, 8*, Article 262. https://doi.org/10.1057/s41599-021-00943-2

Holm, S. (1979). A simple sequentially rejective multiple test procedure. *Scandinavian Journal of Statistics, 6*(2), 65–70. https://doi.org/10.2307/4615733

İstanbullu, S. (2023). Türk kültüründe inançsal, tarihsel ve nesnel bağlamda türkülerde renkler ve anlamları [Colours in folk songs and their meanings in the religious, historical, and objective context in Turkish culture]. *Yegah Müzikoloji Dergisi, 6*(1), 83–106. https://doi.org/10.51576/ymd.1302992

Lindsey, D. T., & Brown, A. M. (2014). The color lexicon of American English. *Journal of Vision, 14*(2), Article 17. https://doi.org/10.1167/14.2.17

Lindsey, D. T., & Brown, A. M. (2021). Lexical color categories. *Annual Review of Vision Science, 7*, 605–631. https://doi.org/10.1146/annurev-vision-093019-112420

Linhares, J. M. M., Pinto, P. D., & Nascimento, S. M. C. (2008). The number of discernible colors in natural scenes. *Journal of the Optical Society of America A, 25*(12), 2918–2924. https://doi.org/10.1364/JOSAA.25.002918

Liu, Z., & Nikitina, L. (2025, October 23–24). *Colouring emotions: A corpus-assisted psycholinguistic study of popular song lyrics* [Conference presentation]. INTRAC 2025, University of Malaya, Kuala Lumpur, Malaysia. https://www.youtube.com/watch?v=268rwlGSg2w

McManus, I. C. (1983). Basic colour terms in literature. *Language and Speech, 26*(3), 247–252. https://doi.org/10.1177/002383098302600305

Mohr, C., Jonauskaite, D., Dan-Glauser, E. S., Uusküla, M., & Dael, N. (2018). Unifying research on colour and emotion: Time for a cross-cultural survey on emotion associations with colour terms. In L. W. MacDonald, C. P. Biggam, & G. V. Paramei (Eds.), *Progress in colour studies: Cognition, language, and beyond* (pp. 209–222). John Benjamins.

Munsell, A. H. (1912). A pigment color system and notation. *The American Journal of Psychology, 23*(2), 236–244. https://doi.org/10.2307/1412843

Nikitina, L. (2024). Fragrance lexicon for analysis of consumer-generated perfume reviews in Russian and English. *MethodsX, 12*, Article 102627. https://doi.org/10.1016/j.mex.2024.102627

Ou, L. C., Luo, M. R., Woodcock, A., & Wright, A. (2004). A study of colour emotion and colour preference. Part I: Colour emotions for single colours. *Color Research & Application, 29*(3), 232–240. https://doi.org/10.1002/col.20010

Özşahin, M. (2023). Söz varlığı bağlamında Başkurt gelin türküleri: Sĕñlevler [Bashkir bride ballads in the context of vocabulary: Sĕñlevs]. *Türkiyat Mecmuası, 33*(1), 259–296. https://doi.org/10.26650/iuturkiyat.1243930

Pennebaker, J. W., Boyd, R. L., Booth, R. J., Ashokkumar, A., & Francis, M. E. (2022). *Linguistic Inquiry and Word Count: LIWC-22* [Computer software]. Pennebaker Conglomerates. https://www.liwc.app

Salih Hassan, H. (2000). Symbolism of purple in Emily Dickinson's poetry. *Adab Al-Rafidayn, 43*(67), 43–74. https://doi.org/10.33899/radab.2013.83214

Ševčíková, M. (2008). *Symbolism of white in the poetry of Emily Dickinson* [Master's thesis, Masaryk University]. Masaryk University Information System. https://is.muni.cz/th/aj18q/Symbolism_of_White_in_the_Poetry_of_Emily_Dickinson.pdf

Umam, K., & Nirmala, D. (2020). Colour metaphor in English and Indonesian song lyrics. *RETORIKA: Jurnal Bahasa, Sastra, dan Pengajarannya, 13*(1), 66–72. https://doi.org/10.26858/retorika.v13i1.11504

Valdez, P., & Mehrabian, A. (1994). Effects of color on emotions. *Journal of Experimental Psychology: General, 123*(4), 394–409. https://doi.org/10.1037/0096-3445.123.4.394

Wierzbicka, A. (2008). Why there are no colour universals in language and thought. *Journal of the Royal Anthropological Institute, 14*(2), 407–425. https://doi.org/10.1111/j.1467-9655.2008.00509.x